\documentclass[acmtog]{acmart}

\usepackage{wrapfigSafe}
\usepackage{nicefrac}
\usepackage{mathtools}
\usepackage{graphicx}
\usepackage{booktabs} 
\usepackage{combelow} 
\usepackage[ruled,vlined,linesnumbered]{algorithm2e}
\usepackage{tabularx} 
\usepackage{colortbl} 
\usepackage{cancel}
\usepackage{makecell} 
\usepackage{enumitem}
\usepackage{bm}
\usepackage{mathrsfs}

\usepackage{manfnt} 
\usepackage[nameinlink]{cleveref}
\usepackage{multirow}
\usepackage{diagbox}
\usepackage{booktabs}
\usepackage{adjustbox}
\usepackage{array}
\usepackage[usestackEOL]{stackengine}
\usepackage{cuted}
\usepackage{capt-of}
\usepackage{enumitem,kantlipsum}
\usepackage{ccicons}
\usepackage{amsmath}
\usepackage{amsfonts}
\usepackage{amsbsy}
\usepackage[normalem]{ulem} 

\newcolumntype{C}[1]{>{\centering\arraybackslash}p{#1}}

\usepackage{listings} 
\usepackage{courier}
\definecolor{mygreen}{rgb}{0,0.6,0}
\definecolor{mygray}{rgb}{0.5,0.5,0.5}
\definecolor{mymauve}{rgb}{0.58,0,0.82}
\definecolor{superlightgray}{RGB}{240,240,240}
\definecolor{darkBlue}{RGB}{0, 94, 184}
\definecolor{derekBlue}{RGB}{144,210,236}
\definecolor{derekTableBlue}{RGB}{189,235,252}
\definecolor{iglGreen}{RGB}{153,203,67}
\definecolor{coralRed}{RGB}{250,114,104}
\definecolor{gray}{RGB}{180,180,180}
\definecolor{orange}{RGB}{255,165,0}
\definecolor{TechnionBlue}{RGB}{8,33,78}
\definecolor{Purple}{RGB}{137, 99, 198}
\definecolor{lightgray}{gray}{0.65}

\newif\ifshownotes
\shownotestrue 

\ifdefined\MakeWithNotes
  \shownotestrue
\fi
\ifdefined\MakeWithoutNotes
  \shownotesfalse
\fi

\ifshownotes
  \newcommand{\colornote}[3]{{\color{#1}\bf{#2: #3}\normalfont}}
  \newcommand{\colornoteTwo}[3]{{\color{#1}\bf{#3}\normalfont}}
  \newcommand{\colornoteThree}[2]{{\color{#1}\bf{#2}\normalfont}}      
\else
  \newcommand{\colornote}[3]{}
  \newcommand{\colornoteTwo}[3]{}
  \newcommand{\colornoteThree}[2]{}      
\fi

\definecolor{darkgreen}{rgb}{0.0,0.65,0}

\SetCommentSty{commentFont}
\SetNlSty{textbf}{}{.}

\newcommand*{\reffig}[1]{%
  \begingroup
    \def\figureautorefname{Fig.}%
    \autoref{fig:#1}%
  \endgroup
}

\def\MethodNameUp{Kinematic Kitbashing}
\def\MethodNameLow{kinematic kitbashing}

\newcommand{\vecFont}[1]{\mathbf{#1}}

\def\vd{{\vecFont{d}}}

\def\vu{{\vecFont{u}}}

\def\vx{{\vecFont{x}}}

\newcommand{\matFont}[1]{\mathbf{#1}}

\def\mF{{\matFont{F}}}

\def\mP{{\matFont{P}}}
\def\mQ{{\matFont{Q}}}
\def\mR{{\matFont{R}}}

\def\mT{{\matFont{T}}}

\def\mV{{\matFont{V}}}

\newcommand{\R}{\mathbb{R}}
\newcommand{\M}{\mathcal{M}}

\newcommand{\V}{\mV}

\newcommand{\vdf}{\vu}
\newcommand{\A}{\mathcal{A}}

\DeclareMathOperator{\Exp}{Exp}
\DeclareMathOperator{\Log}{Log}

\newcommand{\jointq}{{\bm{\theta}}}
\newcommand{\jointT}{\mT}

\newcommand{\Sim}{\mathrm{Sim}(3)}

\newcommand{\optP}{\mP}
\newcommand{\optPConcat}{\mathcal{P}}

\newcommand{\artShape}{\mathcal{S}}
\newcommand{\artPlacedShape}{\widetilde{\mathcal{S}}}

\newcommand{\Ekm}{E^{\mathrm{km}}}
\newcommand{\Efunc}{E^{\mathrm{func}}}

\newcommand{\MParent}{\widebar{\M}} 
\newcommand{\MParentNew}{\widebar{\M}^{\mathrm{new}}}

\newcommand{\artMesh}{\mathcal{A}}
\newcommand{\artPlacedMesh}{\widetilde{\mathcal{A}}}

\AtBeginDocument{%
  \providecommand\BibTeX{{%
    \normalfont B\kern-0.5em{\scshape i\kern-0.25em b}\kern-0.8em\TeX}}}

\setcopyright{acmcopyright}
\acmDOI{0000001.0000001_2}

\copyrightyear{2026}
\acmYear{2026}
\setcopyright{cc}
\setcctype{by}
\acmConference[SIGGRAPH Conference Papers '26]{Special Interest Group on Computer Graphics and Interactive Techniques Conference Conference Papers}{July 19--23, 2026}{Los Angeles, CA, USA}
\acmBooktitle{Special Interest Group on Computer Graphics and Interactive Techniques Conference Conference Papers (SIGGRAPH Conference Papers '26), July 19--23, 2026, Los Angeles, CA, USA}
\acmDOI{10.1145/3799902.3811162}
\acmISBN{979-8-4007-2554-8/2026/07}

\acmSubmissionID{946}

\setcitestyle{square} 

\begin{document}

\title{Kinematic Kitbashing}

\author{Minghao Guo}
\affiliation{%
  \institution{MIT, CSAIL}
  \country{USA}
  }
\author{Victor Zordan}
\affiliation{%
  \institution{Roblox, Clemson University}
  \country{USA}
  }
\author{Sheldon Andrews}
\affiliation{%
  \institution{École de technologie supérieure}
  \country{Canada}
  }
\author{Wojciech Matusik}
\affiliation{%
  \institution{MIT, CSAIL}
  \country{USA}
  }
\author{Maneesh Agrawala}
\affiliation{%
  \institution{Stanford University}
  \country{USA}
  }
\author{Hsueh-Ti Derek Liu}
\affiliation{%
  \institution{Roblox}
  \country{Canada}
  }
  


\renewcommand{\shortauthors}{Guo et al.}

\newcommand{\mh}[1]{\textcolor{red}{[MG: #1]}}
\newcommand{\fr}[1]{#1}
\newcommand{\frm}[1]{#1}

\begin{strip}\centering
\includegraphics[width=\textwidth]{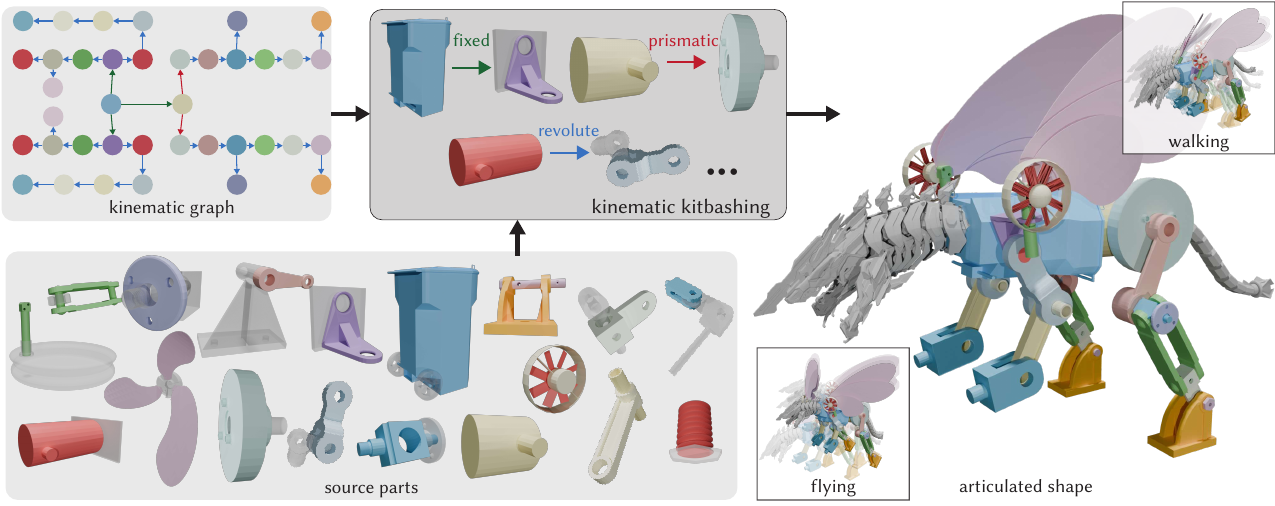}
\vspace{-0.7cm}
\captionof{figure}{\textbf{We introduce {\MethodNameUp}, a framework that synthesizes articulated objects by reusing parts from existing models.} Our method instantiates an abstract kinematic graph with exemplar-based attachment cues to determine the precise placement and scale of each part. The result is a fully realized articulated model (illustrated by this mechanical dragon) that supports interactive behaviors like walking and flight.
}
\vspace{-0.1cm}
\label{fig:teaser}
\end{strip}

\begin{abstract}

We introduce Kinematic Kitbashing, an optimization framework that synthesizes articulated 3D objects by assembling reusable parts conditioned on an abstract kinematic graph. Given the graph and a library of articulated parts, our method optimizes per-part similarity transformations that place, orient, and scale each component into a coherent articulated object; optional graph edits further enable novel assemblies beyond the prescribed connectivity. Central to our method is an exemplar-based analogy for part placement: each reused component is paired with a single source asset that exemplifies how it attaches to its parent. We capture this attachment context using vector distance fields and measure consistency by integrating the matching error over the joint’s full motion range. This yields a kinematics-aware attachment energy that favors placements that preserve the exemplar’s local attachment neighborhood throughout articulation.  To incorporate task-level functionality, we use this attachment energy as a prior in an annealed Langevin sampling framework, enabling gradient-free optimization of black-box functionality objectives. We demonstrate the versatility of kinematic kitbashing across diverse applications, including instantiating kinematic graphs from user-selected or automatically retrieved parts, synthesizing assemblies with user-defined functionality, and re-targeting articulations via graph edits. 

\end{abstract}




\maketitle

\section{Introduction}


Creating articulated 3D shapes that adhere to a functional \emph{schema} is essential for interactive virtual environments, from drivable vehicles to characters with rich animation. 
An \emph{articulation} specifies the kinematic degrees of freedom (DoFs) of a multi-part assembly, while a \emph{schema} specifies how these DoFs are exposed to external control.
For instance, a vehicle schema may map throttle to wheel rotations or link a discrete key-press to a locking state that governs a door joint's mobility. 
When assets conform to a shared schema, developers can reuse the same interaction logic across many geometric variants, enabling rapid content creation without re-authoring behaviors.
%
%

Yet, authoring such assets remains difficult: one must simultaneously satisfy coupled constraints of geometry, articulation, and schema.
This often necessitates labor-intensive manual modeling or specialized interfaces tailored to a particular schema.
While prior work has advanced articulated object creation, many methods neglect explicit schema compliance, often producing assets that support only limited, passive physical interactions. 
%
This gap motivates our method, which integrates schema constraints directly into articulated object synthesis.

The structural core of an articulation schema is its \emph{kinematic graph}, where nodes represent rigid parts and edges represent joints. 
In practice, this graph is often defined abstractly at the level of part-joint connectivity: 
it captures which parts are linked, but not their precise 3D configurations, orientations, or scales. 
This abstraction enables reuse across asset families: many objects share the same part-joint connectivity despite having vastly different geometries. 
For instance, a poseable character schema typically connects a torso to multiple limb and head chains, yet can represent a wide range of creatures independent of the specific mesh geometry (see \reffig{teaser}).


In this work, we focus on the \emph{geometric instantiation} of an abstract kinematic graph into a fully functional articulated object that complies with a given schema.
We take inspiration from \emph{kitbashing}\footnote{\url{https://en.wikipedia.org/wiki/Kitbashing}}, the practice of reusing components from existing models for novel assemblies. 
We extend this idea from static shapes to articulated objects by casting geometric instantiation as an exemplar-based analogy: for each reused child part, a single source asset provides an exemplar of how that part should attach to a new parent.
The exemplar is informative because the parent geometry around the child typically forms a characteristic neighborhood, such as a socket, recess, or clearance region, that accommodates the child's intended motion.
We treat this local geometric affordance as a transferable cue for reattachment;
for example, a wheel is typically seated in a hub-like recess providing support and clearance for rotation. 

With this insight, we capture attachment cues with a vector distance field (VDF) that measures offsets from a part to its surrounding parent.
Because an articulation spans a continuum of poses, we accumulate the VDF-matching error across the entire pose dimension, yielding a \emph{kinematics-aware} energy that favors placements whose surrounding geometry remains consistent with the exemplar throughout the entire articulation cycle. 
Aggregating over poses effectively adds a \emph{time} dimension on top of the \emph{spatial} dimension. 
We address this added complexity with an alternating optimization scheme that keeps runtime practical.

Our \emph{\MethodNameLow} framework supports several workflows for articulated object creation. 
When a kinematic graph is prescribed, our kinematics-aware attachment directly instantiates it from user-selected or retrieved parts, enabling both kinematics-conditioned assembly and mix-and-match synthesis. 
Beyond kinematic structure, assets often must satisfy task-level functional goals, such as collision-free actuation, reachability, or trajectory tracking, which are typically evaluated through black-box procedures (e.g., inverse kinematics or collision queries). 
We incorporate these heterogeneous objectives within an annealed Langevin sampling framework, using our attachment energy as a strong prior to focus global exploration on geometrically plausible assemblies. 
The framework further supports articulation re-targeting by exploring modest graph edits to discover variants that improve functional performance while maintaining high-quality attachments. 

In summary, we make the following contributions: 
(1) an exemplar-based kinematic kitbashing formulation for instantiating kinematic graphs by reusing articulated parts; 
(2) a kinematics-aware VDF attachment energy that aggregates geometric matching errors across the full articulation cycle; 
and (3) a general optimization framework that combines this attachment energy with black-box functional objectives and supports kinematics-conditioned assembly, functionality-guided synthesis, and articulation re-targeting.

\section{Related Work}
%
%
%
We focus on synthesizing \emph{functional}, \emph{articulated} objects, and refer readers to recent surveys 
\cite{MitraWZCKH13, ChaudhuriRWXZ20, RitchieGJMSWW23, XuGFCR23} for static object synthesis.

\subsection{Articulated Mechanical Assemblies} \label{sec:related_assembly}
The modeling of \emph{articulated mechanical assemblies}
has become a topic of significant interest for the computer graphics and computational fabrication communities. 
Prior work has focused chiefly on semi-automatic and optimization-based workflows for creating such assemblies, with applications in robotics, animation, and computational fabrication. 
Methods for physically fabricable mechanisms include complex linkages~\cite{CorosTNSFSMB13,Thomaszewski2014}, compliant joints~\cite{Megaro2017b, Zhang2021}, and cable-driven systems~\cite{Megaro2017, Li2017}.
Other approaches focus on robotic characters with expressive or task-specific motions~\cite{Megaro2015,Maloisel2023,Geilinger2018,ChaCra2015,Ceylan2013}, and dynamic objects~\cite{SorkineHornung14}.

A common strategy across many of these methods is to begin with a high-level specification of motion, such as a reference trajectory~\cite{Thomaszewski2014, CorosTNSFSMB13} or an animation sequence~\cite{Ceylan2013}, and then solve a non-linear optimization problem to determine the parameters of the mechanical assembly that best realizes these motions. 
While powerful, these methods require a carefully designed differentiated forward model and are usually tailored to a particular mechanism class; adapting them to new objectives often requires substantial re-engineering. 
Our approach, by contrast, 
accepts black-box functional objectives. 
This offers a practical and scalable alternative to existing methods, particularly in cases where rapid exploration of designs is essential.

\subsection{Modeling Articulated Objects}\label{sec:articulated_modeling}
A line of research focuses on modeling \emph{articulated objects}, i.e., a collection of rigid body components connected by mechanical joints, with the aim of producing visually plausible articulations\,\cite{liu2024survey}.
One approach is to distill the articulations between the parts using images/videos\,\cite{pekelny2008articulated, SongWFLL24, wu2025predict, Ditto, nie2023structure, PARIS, YaoHLR0J22} or from pre-trained language/vision models \cite{SINGAPO, ArticulateAnything, mandi2024real2code, vora2025articulateobjectatop3d, artiWorld, FreeArt3D}. 
When having access to ground truth articulation datasets (e.g., PartNetMobility \cite{PartNetMobility} or procedural generation \cite{joshi2025infinigensimproceduralgenerationarticulated}), several works have demonstrated successfully infer articulation parameters for an input object made up of parts \cite{WangZSCZ019, YanHYCKZH19, FuISO24, PhysPart, goyal2025geopardgeometricpretrainingarticulation, S2O}. 
These works, however, assume that the input parts are already correctly arranged with respect to each other, which is inapplicable to our problem with unknown arrangements of input geometric parts.
The works closest to ours are the methods of \citet{CAGE} and \citet{NAP}. From the given kinematic graph connectivity, they use graph denoising diffusion to generate parts and articulation parameters jointly. 
\citet{SuFLSHRX25, GaoSLD25} take a step further to also generate the kinematic graphs with an autoregressive model, followed by parts generation. 
However, these methods depend on training neural networks on available datasets, which are limited in size and restrict their ability to handle configurations beyond the training distribution. 
Our approach complements these generative models  by providing a ``one-shot'' solution that refines articulated objects from single exemplars. 
As shown in Fig.~\ref{fig:storage_edit}, our method can directly refine the articulated objects generated by \citet{FreeArt3D}.

\begin{figure}[t]
    \begin{center}
    \includegraphics[width=1\linewidth]{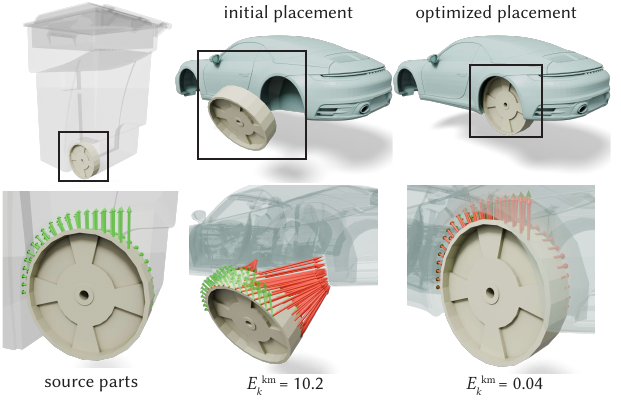}
    \end{center}
    \vspace{-3ex}
    \caption{\textbf{Kinematics-aware attachment for a novel part pair.} A wheel mounted on a trash can is reattached to a car body. Green vectors show the exemplar VDF from the wheel to its original parent (trash can), and red vectors depict the current VDF to the new parent (car body) under the current placement. Our VDF energy guides the wheel to the fender recess, matching its original socket. 
    }
    \vspace{-1ex}
    \label{fig:attachment-snapping}
\end{figure}

\subsection{Synthesis of Functional Objects} \label{sec:related_functional_synthesis}
{\em Topology optimization}\,\cite{deaton2014survey} is a broad term for work that considers functionality in 3D shape modeling.  Work in this area typically treats structural properties (e.g., material distributions) of the object as a variable that is optimized to satisfy functional goals (e.g., minimizing compliance given a target load). 
However, these methods primarily optimize the interiors of single objects, contrasting with our focus on the relationship between discrete articulated components. 
Another relevant area is \emph{functional shape analysis}\,\cite{HuSK18}, which encompasses visualizing and characterizing functional objects. 
\fr{
Part-based generative methods such as GRASS~\cite{li2017grass} and Fit and Diverse~\cite{xu2012fit} also synthesize diverse shapes by modeling or exploring structural part arrangements; however, they target static shape plausibility and diversity rather than motion-dependent attachments between articulated components.
}
Closer to our approach, \citet{ZhengCM13} characterize functionality through component arrangements. 
By preserving simple geometric relationships (e.g., symmetries and contacts), the method enables mix-and-match synthesis of static objects with similar functionality. 
%
\citet{JacobsonG16} propose an interactive interface to allow users to monitor and resolve infeasible configurations.
\citet{YaoKGA17} present an interactive tool for creating interlocking woodworking-style joints for furniture and toys, along with tools for verifying the object's overall ease of assembly and stability.
%
\citet{BesiegeField} explore agentic workflows on a virtual playground with simulation feedback to synthesize functional objects.
\citet{LiAM14} optimize the arrangement of geometric proxies (specifically cuboids), based on user-specified relationships, such as \emph{coverage}, \emph{fit inside}, \emph{support}, and \emph{flush}.
Our work 
generalizes
the method by 
\cite{LiAM14} to triangle meshes and incorporates global functional goals (e.g., reachability across multiple articulation poses) that depend on the configuration of the entire assembly.

\section{Approach}
 %

We frame \emph{kinematic kitbashing} as an optimization that places a given set of articulated parts into a shared world frame.
Specifically, the input consists of 
(1) a kinematic graph whose nodes represent a set of kinematic parts, and edges represent their parent-child adjacency in the articulation,
and (2) a set of kinematic parts sourced from a database of articulated assets.
Each part is sourced from a single exemplar asset, which provides the part's rest-pose mesh, its child-side joint parameters, and the geometry of its original parent.
This original parent-child pair serves as the exemplar attachment for reattaching the part to a new parent.
We do not assume access to the corresponding joints on the parent geometry. 
Instead, the optimization searches for the similarity transformations required to place the parts such that they satisfy the kinematic graph while optionally maximizing a functionality objective (e.g., reachability or trajectory tracking).
This setup reflects realistic scenarios for articulated object modeling:
Parts are straightforward to collect, but manually annotating precise attachment positions and part poses on every potential parent mesh for desired functional goals is both tedious and error-prone. 

\subsection{Problem Formulation \& Notations}
An articulated object consists of a set of rigid parts linked by mechanical joints that constrain their relative motion~\citep{featherstone2014rigid}.
The joint connectivity forms a directed graph. 
We represent each rigid part together with the joint that connects it to its parent as a \emph{kinematic part}.

Let $\artMesh_k=(\mathbf{V}_k, \mathbf{F}_k)$ be the mesh of the $k$-th kinematic part in the graph, where $\mathbf{V}_k$ and $\mathbf{F}_k$ denote its vertices and faces.
Each joint is parameterized by a generalized coordinate $\theta\!\in\![\theta_\text{lo}, \theta_\text{hi}]$, whose dimension equals the DoFs of the articulated motion.
The joint parameters of all the parts are collected into $\jointq_i$, which specifies an articulation pose $i$ of the articulated object.
\emph{Forward kinematics} function $\jointT(\V_k; \jointq_i)\!:\!\R^{3\times |\V_k|}\times\R^{d}\!\to\!\R^{3\times |\V_k|}$ with $d$ as the total number of DoFs of all joints,
propagates these joint parameters $\jointq_i$ along the kinematic graph and outputs a set of transformed vertex locations from the rest-pose vertices $\V_k$.
We write the $k$-th part $\A^{(i)}_k$ at an
articulated pose indexed by $i$ as 
\begin{align}
    \artMesh^{(i)}_k = \big(\ \jointT(\V_k; \jointq_i), \mF_k\ \big) \,.
\end{align}
An articulated object $\artShape$ in reference coordinates consists of a set of kinematic parts at different articulation poses:
\begin{align}
    \artShape = \{ \A^{(i)}_k \} \,.
\end{align}

\begin{figure}
    \begin{center}
    \includegraphics[width=1\linewidth]{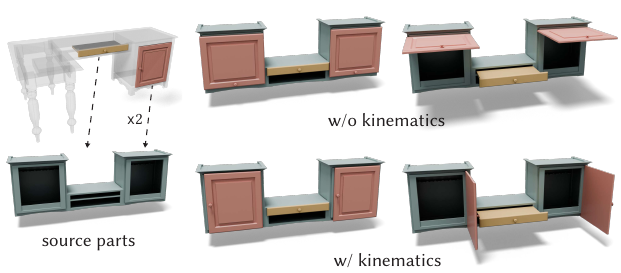}
    \end{center}
    \vspace{-3ex}
    \caption{\textbf{Effect of kinematic attachment.} A drawer slide and a hinged door taken from the table are attached to a cabinet. Without kinematic-aware attachment (top), i.e., the placement is optimized for a single pose and ignores the full articulation, the doors are misoriented.
    With the kinematic dimension (bottom), our method positions both parts plausibly.}
    \vspace{-1ex}
    \label{fig:attachment-comparison-with-prior}
\end{figure}

Given a collection of kinematic parts sourced from exemplar articulated objects, our kinematic kitbashing aims to compute a set of orientation-preserving similarity transformations 
%
\begin{align}
    \optPConcat = \big\{\optP_k\big\}_{k=1}^K = \big\{\ (\mR_k, \vd_k, s_k)\ \big\}_{k=1}^K
\end{align}
\frm{
\begin{wrapfigure}[9]{r}{1.67in}
    \vspace{-8pt} 
    \includegraphics[width=\linewidth, trim={5mm 0mm 0mm 0mm}]{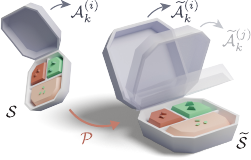}
\end{wrapfigure}
}
to place each part $\artMesh^{(i)}_k$ in world coordinates.
We use $\mR_k\!\in\!\mathrm{SO}(3)$, $\vd_k\!\in\!\mathbb{R}^3$, and $s_k\!\in\!\mathbb{R}^{+}$ to denote rotation, translation, and scaling, respectively, for part $k$. $\optP_k$ denotes the homogeneous similarity transformation matrix formed from $\mR_k, \vd_k, s_k$.
We use $\artPlacedMesh^{(i)}_k(\optPConcat)$ to denote the 
assembled instance
of $\artMesh^{(i)}_k$ once it has been positioned in world coordinates (an illustrative example in the inset):
\begin{align}
    \artPlacedMesh^{(i)}_k(\optPConcat) = \big(\ \jointT(s_k\mR_k \V_k + \vd_k; \jointq_i), \mF_k\ \big).
\end{align}
Similar to the usage of $\artShape$ for the articulated shape in reference coordinates, we use  
\begin{align}
    \artPlacedShape(\optPConcat) = \{\ \artPlacedMesh^{(i)}_k(\optPConcat) \ \}
\end{align}
to denote the collection of all articulated components across all poses in world coordinates, determined by the set of transformations $\optPConcat$.

\subsection{Kinematics-Aware Geometric Attachment}
\label{sec:vdf-attachment}
The kinematic attachment term $\Ekm$ is designed to support assemblies with complex kinematic graphs (\reffig{teaser}) and cross-category parts (\reffig{attachment-snapping}).
Our key insight is that articulation exposes kinematic features in the geometry of the parent part, specifically in the region that is spatially surrounding the kinematic part. 
This geometric region encodes cues that should be preserved when reattaching the part to a new parent. 
For example, the wheel on a trash can rotates relative to the container, and the surrounding geometry on the container resembles a hub slot that accommodates this rotation. 
These features provide a strong geometric guide for attaching parts: when snapping the wheel to a new parent, such as a car body, it should be placed at a location that exhibits similar geometric affordances. 

\subsubsection{Vector Distance Field Attachment Energy}
We encode these surrounding region cues using a \emph{vector distance field} (VDF)~\cite{vectorDistanceFunction, neuralpull, MarschnerSLJ23}.
Given a source mesh $\M^s$ and a reference mesh $\M^t$, the VDF assigns each vertex $\vx \in \M^s$  a vector from $\vx$ to its closest point on $\M^t$, such that
\begin{align}\label{equ:vdf}
    \vdf(\M^s;\M^t) = \{\mathrm{proj}_{\M^t}(\vx) - \vx |\vx \in \M^s\} \,,
\end{align}
where $\mathrm{proj}_{\M^t}(\vx)$ is the projection operator. 
Throughout this paper, we adopt the sign convention that the vectors point toward the surface of the reference mesh.

%

%

%

%
\frm{
\begin{wrapfigure}[8]{r}{1.8in}
    \vspace{-10pt} 
    \includegraphics[width=\linewidth, trim={5mm 0mm 0mm 0mm}]{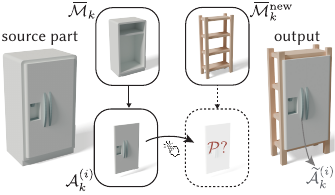}
\end{wrapfigure}
}
Let $\MParent_k$ be the parent mesh of $\artMesh^{(i)}_k$ in a source articulated object. 
For every part-parent pair $\vdf(\artMesh^{(i)}_k, \MParent_k)$,
we compute the VDF 
by projecting the surface vertices of the part $\artMesh^{(i)}_k$ onto its parent mesh $\MParent_k$ over a set of articulation parameters $\{\jointq_i\}$ uniformly sampled over the joint range.
%
%
At attachment time, given a new parent part $\MParentNew_k$, we optimize placement transformations $\{ \mR_k, \vd_k, s_k \}$ so that the VDF between the transformed part $\artPlacedMesh^{(i)}_k$ and its new parent $\MParentNew_k$ matches the VDF observed in the source pair $(\artMesh^{(i)}_k, \MParent_k)$.
%
The VDF matching energy for a part $k$ is defined as:
\begin{align}\label{equ:Ekm}
    \Ekm_k\!=\!\sum_{i=1}^{N} \Ekm_{k,i}\!=\!\sum_{i=1}^N \Big\|  s_k\mR_k \vdf\Big(\mathcal{A}_{k}^{(i)};\MParent_k\Big)\!-\!\vdf\Big(\artPlacedMesh^{(i)}_k(\optPConcat); \MParentNew_k\Big) \Big\|^2. 
\end{align}
%
Here $\Ekm_k$ measures each child-parent pair in the kinematic graph and then sums over the $N$ sampled articulated poses indexed by $i$, where each pose is specified by the joint parameters $\jointq_i$ sampled over the joint parameter range.
Aggregating the VDF error across the entire pose range makes the energy \emph{kinematics-aware}: the placement must align the functional neighborhood of the part not just at a single pose but consistently throughout its articulation.
The total $\Ekm \!=\! \sum_k \Ekm_k$ is minimized with respect to the set of rigid placement $\optPConcat=\{(\mR_k,\vd_k, s_k)\}$ across all part pairs and poses.

\begin{figure}
    \begin{center}
    \includegraphics[width=1\linewidth]{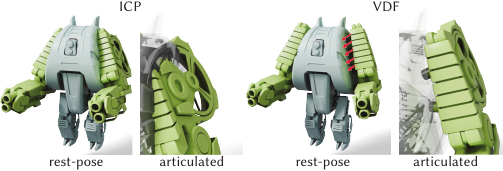}
    \end{center}
    \vspace{-3ex}
    \caption{\textbf{ICP vs. VDF attachment.} ICP ignores the clearance required between parts and snaps the arm directly into the torso surface. When the arm articulates, this oversight becomes critical, producing collisions. VDF, in contrast, keeps the necessary gap (red arrows) for collision-free articulation.
    }
    \label{fig:icp-vs-vdf}
\end{figure}

We minimize $\Ekm_k$ iteratively:
At each iteration, we first recompute VDF vectors between the articulated shape and its new parent $\vdf(\artPlacedMesh^{(i)}_k, \MParentNew_k)$, and then update the similarity transform $(\mR_k, \vd_k, s_k)$ to reduce the residuals. To enhance robustness, we incorporate two techniques inspired by the iterative closest point (ICP) algorithm:  (1) We compute point-to-plane alignment to allow tangential sliding~\cite{chen1992object}. (2) We also adopt Welsch’s weighting function on the residuals to suppress outliers~\cite {sorkine2017least,holland1977robust}.
Our complete formulation is provided in the Appendix. 
Compared to ICP, VDFs inherently respect the clearance between a part and its parent,
whereas ICP’s nearest-point matches drive the surfaces into direct contact and ignore the gap needed for articulation, as shown in Fig.~\ref{fig:icp-vs-vdf}.

\subsubsection{Alternating Local-Global Optimization}
Although the attachment energy $\Ekm$ can, in principle, be optimized using standard second-order methods, we observe that these approaches become computationally expensive due to the integration over the joint parameters $\jointq_i$ at all articulation poses. 
Evaluating gradients across many articulation snapshots increases runtime significantly, especially when fine temporal resolution is required.
To improve performance, we adopt an alternating optimization approach inspired by~\citet{bouaziz2014projective},
instead of a Newton-type solver.
Rather than optimizing one transform that must fit \emph{all} articulation poses simultaneously, we assign each pose $i$
an auxiliary transformation $\mQ_i$ and bind these transforms to a shared similarity transformation $\optP_k \!=\! (\mR_k,\vd_k, s_k)$ for part $k$.
The subproblems for computing $\mQ_i$ are now independent and can be solved in parallel, and $\optP_k$ can be updated in closed form as the Lie-algebra average of all $\mQ_i$.
This decoupling greatly reduces the runtime.
Specifically, 
we optimize the following modified objective:
\begin{alignat}{1}
    \mathrm{Local \ Step:}\quad &\min_{\mQ_i} \sum_{i=1}^{N} \Bigl(\Ekm_{k,i} + \tfrac{\rho}{2}\bigl\lVert \Log\bigl(\mQ_i^{-1}\optP_k \bigr)\bigr\rVert^{2}\Bigr) \,, \\[-3pt]
    \mathrm{Global \ Step:} \quad &\ \optP_k  = \Exp\Bigl(\frac{1}{N}\sum_{i=1}^{N}\Log \mQ_i\Bigr) \,,
\end{alignat}
%
where $\Log(\cdot)\!:\!\Sim\!\to\!\mathbb R^{7}$ is the vectorized logarithm map in Lie algebra
and $\Exp(\cdot)$ is its inverse map~\cite{chirikjian2011stochastic}. 

\begin{figure}[t]
    \begin{center}
    \includegraphics[width=1\linewidth]{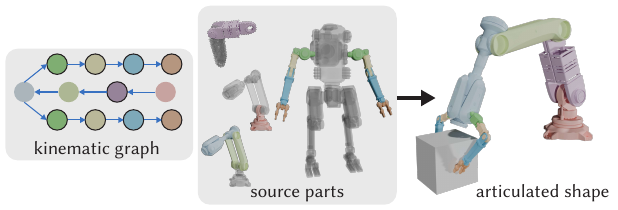}
    \end{center}
    \vspace{-3ex}
    \caption{\textbf{Scale-adaptive kinematics-conditioned assembly.} An articulated arm chain is synthesized by inserting an intermediate link and reusing two arms from a robotic character. Our optimization properly estimates a down-scaling of the reused arms, ensuring their fit to the new attachment geometry while preserving the intended articulation.}
\vspace{-1ex}
    \label{fig:different-priors}
\end{figure}
In the local step, to account for the non-Euclidean nature of $\Sim$, we replace the standard Euclidean penalty with a geodesic distance between $\optP_k$ and $\mQ_i$, defined via the logarithmic map on $\Sim$. 
The global step similarly computes the average of $\mQ_i$ in the Lie algebra and maps it back to the group using $\Exp(\cdot)$, ensuring that $\optP_k$ remains on the $\Sim$ manifold.
We set $\rho\!=\!10^4$ to weight the penalty term.
Optimization proceeds by alternating between the local and the global steps.
We execute up to $30$ iterations.
Compared to Newton-type optimization methods, this approach significantly improves performance by enabling parallel optimization of all $\mQ_i$ in the local step.
For $N\!=\!20$ in the door example (Fig.~\ref{fig:attachment-comparison-with-prior}), our approach reduces the wall-clock optimization time from $163$~s to $19$~s. 
\section{Applications and Results}

We demonstrate the practical scope of our kinematic kitbashing framework through several applications.
All results instantiate a kinematic graph by solving for per-part similarity transformations that place the parts into a shared world frame.
When a functionality objective is provided, such as trajectory-tracking error, swept-volume overlap, or reachability, we optimize the combined objective
\begin{align}
\label{eq:opt-objective}
    \min_{\optPConcat} \quad \Ekm(\artPlacedShape(\optPConcat)) + \lambda \Efunc(\artPlacedShape(\optPConcat)) \, ,
\end{align}
where $\Efunc$ represents potentially non-differentiable functionality.

We organize our results as follows: 
Sec.~\ref{sec:4.1} shows the instantiation of a fixed kinematic schema based solely on geometric attachment, without functionality goals. 
We explore both user-selected parts and automated retrieval from large part libraries. 
Sec.~\ref{sec:4.2} extends this framework by augmenting assembly with task-specific objectives. 
Sec.~\ref{sec:4.3} introduces graph edits to the kinematic graph, enabling articulation re-targeting when the original connectivity is ill-suited to the functional requirements.




\subsection{Kinematics-Conditioned Assembly}
\label{sec:4.1}
In many interactive settings, the kinematic graph serves as a reusable schema (e.g., a character skeleton or a furniture template), and the modeling task is to instantiate this graph with concrete geometry.
In this workflow, the kinematic graph is fixed, and we compute the placement transformations by minimizing $\Ekm$ alone ($\lambda=0$).

\begin{figure*}[t]
    \begin{center}
    \includegraphics[width=0.99\linewidth]{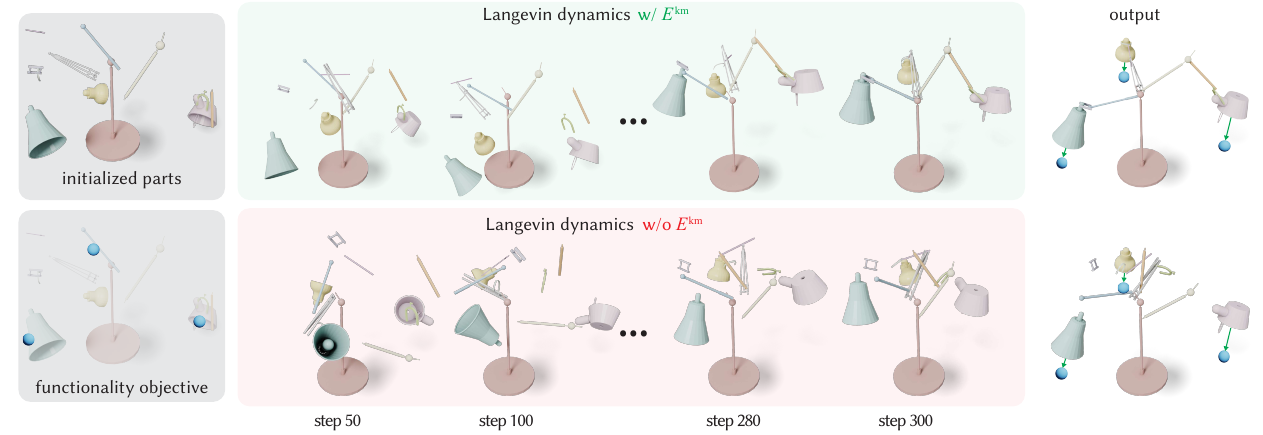}
    \end{center}
    \vspace{-3ex}
    \caption{\textbf{Intermediate results of annealed Langevin sampling.} Langevin dynamics run with both the kinematic attachment term and the functionality term gradually assemble the parts and end with a mechanism whose end-effectors reach the targets (top). Without the kinematic attachment term, the sampler can satisfy the target-reaching objective, but only via geometrically implausible configurations that do not form a meaningful assembly (bottom). }
    \label{fig:langevin-dynamic}
\end{figure*}

\begin{figure}[t]
    \begin{center}
    \includegraphics[width=1\linewidth]{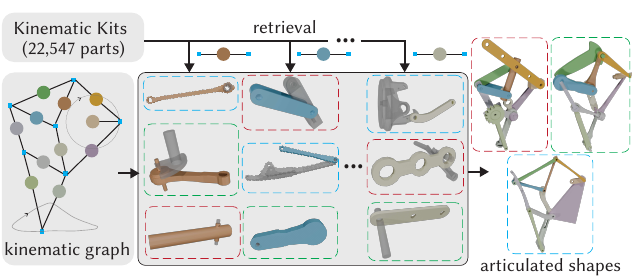}
    \end{center}
    \vspace{-3ex}
    \caption{\textbf{Kinematics-conditioned assembly with part retrieval.} Given the kinematic graph of Jansen’s linkage, we retrieve compatible kinematic parts from a part library and then optimize their transformations in a shared world frame. Different retrieved part sets yield diverse articulated instantiations.}
\vspace{-1ex}
    \label{fig:janson}
\end{figure}

We consider two input settings.
In the first, the user selects parts from the library and assigns them to the nodes of the graph.
This enables graph-constrained mix-and-match synthesis, such as assembling a character that conforms to a predefined kinematic graph (Fig.~\ref{fig:teaser} and~\ref{fig:mix-and-match}), assembling a new robot arm from user-chosen modules (Fig.~\ref{fig:different-priors}), or instantiating storage furniture whose door/drawer layout follows a kinematic schema (Fig.~\ref{fig:expr_baseline_2}).

In the second setting, parts are not provided explicitly and are instead retrieved from a large part library.
Given a graph whose nodes carry coarse geometric cues, e.g., desired oriented bounding-box dimensions and joint locations expressed in the bounding-box frame, we retrieve candidate parts by matching these cues and then run the same kinematics-conditioned assembly optimization.
Fig.~\ref{fig:janson} and~\ref{fig:app_jansen} demonstrate this setting on Jansen's linkage: starting from an unembedded linkage graph with per-node size hints, we retrieve parts from the JoinABLe library~\citep{Joinable} to instantiate multiple geometric assemblies that realize the same articulated mechanism.





\subsection{Functionality-Guided Assembly}
\label{sec:4.2}

\frm{
\begin{wrapfigure}[8]{r}{1.6in}
    \vspace{-18pt} 
    \includegraphics[width=\linewidth, trim={8mm 0mm 0mm 0mm}]{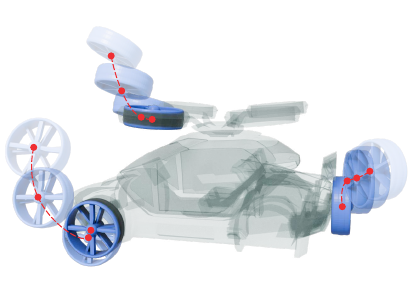}
\end{wrapfigure}
}
Kinematics-conditioned assembly ensures geometrically meaningful attachments, but it may not always uniquely determine a design. 
In practice, minimizing $\Ekm$ alone sometimes admits multiple attachments with similarly low energy, 
\frm{(see inset)}
since $\Ekm$ primarily constrains local parent-child attachment neighborhoods while leaving global behavior underdetermined. 
To accommodate user intent, we 
optimize the combined energy in Eq.~\ref{eq:opt-objective} with $\lambda>0$.
The functionality term $\Efunc$ evaluates task performances such as end-effector reachability, trajectory tracking, collision or swept-volume penalties, and packing constraints. 
For example, in Fig.~\ref{fig:langevin-dynamic}, for a multi-head lamp, we penalize the distance between desired targets and the points illuminated by the lamp heads. 
Since $\Efunc$ is typically evaluated through black-box routines (e.g., inverse kinematics solves and collision queries), we optimize Eq.~\ref{eq:opt-objective} using annealed Langevin sampling (details in the Appendix).
In this joint optimization, $\Ekm$ serves as a strong geometric prior: it concentrates the search within geometrically meaningful attachments. 
This is particularly important for high-DoF assemblies, where assemblies driven solely by $\Efunc$ are often disconnected.
As shown in Fig.~\ref{fig:langevin-dynamic}, incorporating $\Ekm$ leads to geometrically plausible assemblies, whereas removing it results in failure.

\begin{figure}[t]
    \begin{center}
    \includegraphics[width=1\linewidth]{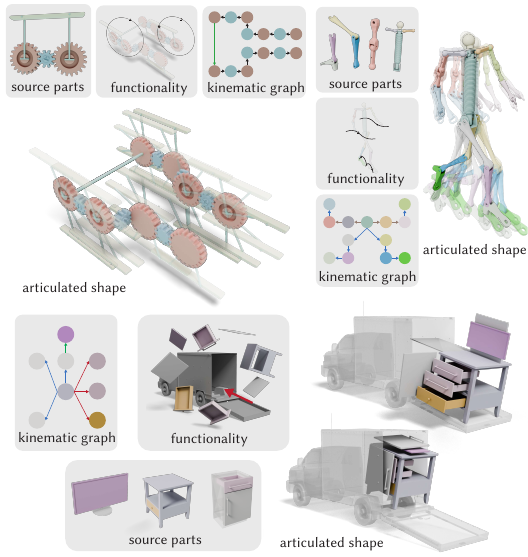}
    \end{center}
    \vspace{-2ex}
    \caption{\textbf{Examples on functionality-guided kinematic kitbashing.} Our method assembles functional mechanisms, including a gear-driven paddle (left), a humanoid composed of mechanical modules (right), and a reconfigurable workstation that folds into a cube to fit inside a truck (bottom).}
    \vspace{-1ex}
    \label{fig:more-examples}
\end{figure}

We demonstrate this workflow on multiple user-specified objectives (Fig.~\ref{fig:more-examples}), including packing a reconfigurable workstation into a compact volume, trajectory tracking for a gear-driven paddle, and a humanoid assembled from mechanical modules.

\subsection{Articulation Re-targeting}
\label{sec:4.3}
In practice, a prescribed kinematic graph is often only a starting point: its connectivity or joint placements may not be well-suited to a new scene or task objective.
Articulation re-targeting extends functionality-guided assembly by allowing the kinematic graph itself to change through discrete edits, while still optimizing continuous part placements.
Starting from an initial graph and part assignment, we explore a small set of candidate graph rewrites (local changes to the parent-child structure and/or joint placement choices), following a similar optimization strategy by~\citet{DesignForDecent}.
For each proposal, we run a short inner optimization of Eq.~\ref{eq:opt-objective} (and, when needed, re-retrieve parts for modified nodes), then select the best-scoring proposal and iterate (details in Appendix).

Fig.~\ref{fig:storage_edit} illustrates this workflow on a cabinet placed in a cluttered environment.
The objective rewards (i) collision-free actuation in the scene and (ii) a closed configuration in which the door panels fully cover the cabinet front.
By proposing graph edits and re-optimizing part placements, the retargeted design achieves both requirements while preserving high-quality kinematic attachments.
A similar graph-editing process is shown in Fig.~\ref{fig:app_jansen}, where we retarget a single-head lamp into a multi-head design by editing the kinematic graph and re-instantiating candidates so that multiple arms can reach the desired tabletop targets.
All examples retrieve parts from PartNet-Mobility~\cite{PartNetMobility}.



\begin{figure}[t]
    \begin{center}
    \includegraphics[width=\linewidth]{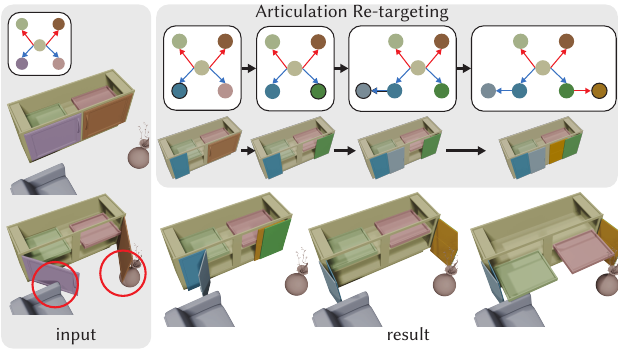}
    \end{center}
    \vspace{-3ex}
    \caption{\textbf{Articulation re-targeting via kinematic graph edits.} Left: an input articulated cabinet placed in a scene; the original articulation leads to collisions with nearby objects. Right: we sample discrete edits to the kinematic graph, re-optimize part placements for each proposal, and score candidates with a functionality objective that encourages both collision-free actuation and a closed configuration in which the door panel fully covers the cabinet front. The re-targeted cabinet opens without collisions while preserving full door coverage (bottom).
}
\vspace{-1ex}
    \label{fig:storage_edit}
\end{figure}

\subsection{Comparison with Baselines}
We benchmark our method against publicly released baselines that share the same inputs as ours to assemble new objects. This includes {4-PCS}~\cite{aiger20084,ChaudhuriK10} and {Part Slot Machine}~\cite{WangGKCSR22} for synthesizing static objects, and {NAP}~\cite{NAP} and {CAGE}~\cite{CAGE} for synthesizing articulated objects conditioned on the kinematic graph. All baseline results are produced with the publicly available code.  
Experiments are run on the \textit{Table} and \textit{Storage Furniture} test split of PartNet-Mobility~\cite{PartNetMobility}, used by CAGE and Part Slot Machine.

Following the baseline papers, we report five metrics: 1) \textit{Rooted}: percentage of assemblies whose parts form a single connected component. 
2) \textit{Stable}: percentage of assemblies that remain upright under gravity.
3) \textit{COV}: coverage, i.e., the fraction of the reference test set that is reproduced by the generated shapes.
4) \textit{MMD}: minimum matching distance between generated and reference test set.
5) \textit{AOR}: average overlapping ratio, the mean volume overlap between any two sibling parts.
These metrics capture geometric (Rooted, Stable), kinematic (COV, MMD), and functional qualities (AOR).
We present quantitative results in Table~\ref{tab:metricWise} and qualitative ones in Fig.~\ref{fig:expr_baseline_2}.
Our method yields the best overall performance.

\begin{figure}[t]
    \begin{center}
    \includegraphics[width=\linewidth]{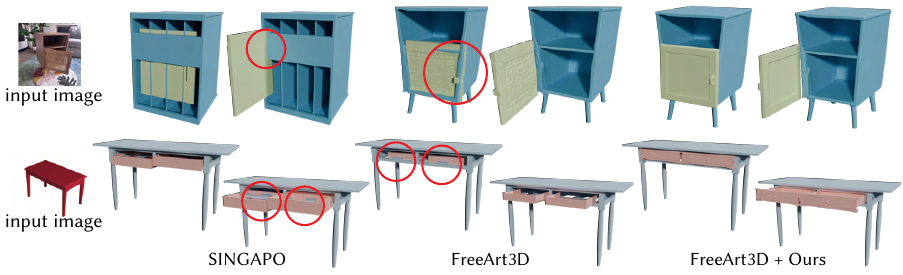}
    \end{center}
    \vspace{-3ex}
\caption{\textbf{Kinematic kitbashing as post-processing for image-based generation.} \fr{SINGAPO~\citep{SINGAPO} and FreeArt3D~\citep{FreeArt3D} generate articulated shapes from a single input image, but  can produce misaligned moving parts and incorrect attachments.} Our kinematic kitbashing {repairs} these outputs by retrieving suitable components and re-instantiating their attachments to yield clean door/drawer alignments.}
\vspace{-1ex}
    \label{fig:post_processing}
\end{figure}

\begin{figure*}[h]
    \begin{center}
    \includegraphics[width=1\linewidth]{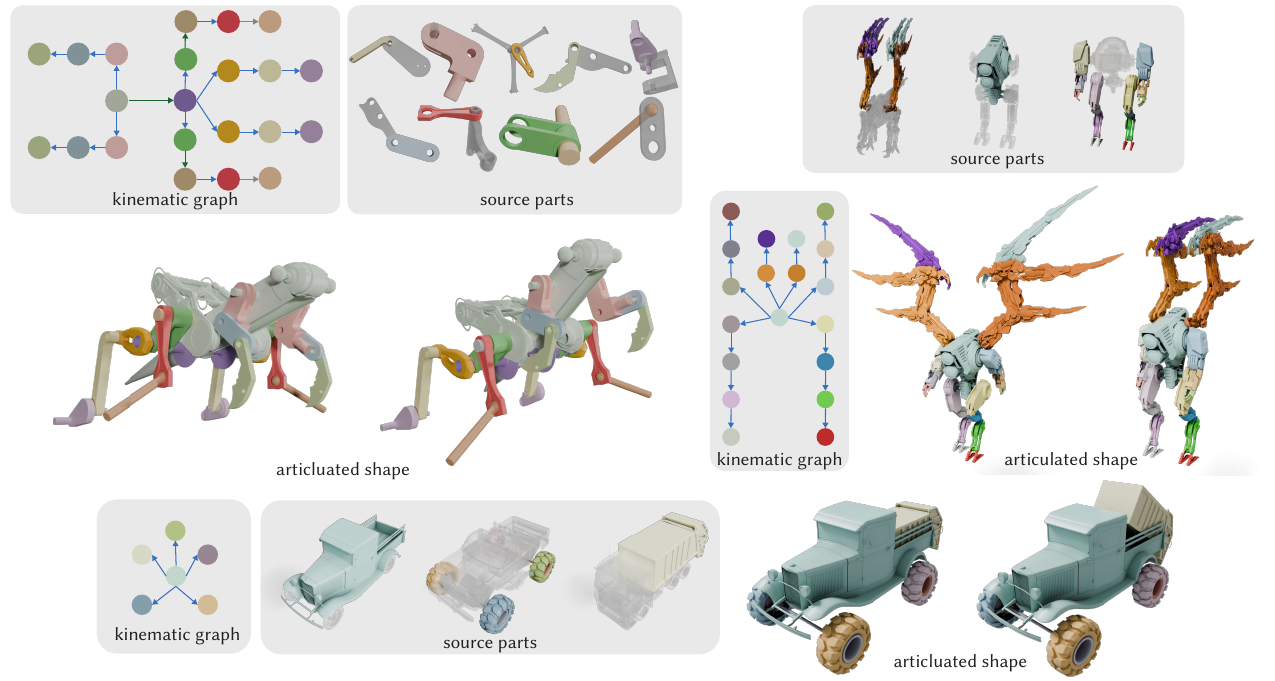}
    \end{center}
    \vspace{-2ex}
    \caption{\textbf{More results on kinematics-conditioned assemblies.} Given a kinematic graph and parts reused from disparate source assets, our kinematics-aware attachment optimizes per-part placements to form a coherent articulated assembly. {Top-left:} a mantis-like mechanical insect assembled from heterogeneous linkage parts.
{Top-right:} robotic legs and wings grafted onto a fantasy torso.
{Bottom:} oversized carriage wheels and a dumpster snapped onto a vintage truck.}
    \label{fig:mix-and-match}
\end{figure*}

\begin{figure*}[h]
    \begin{center}
    \includegraphics[width=1\linewidth]{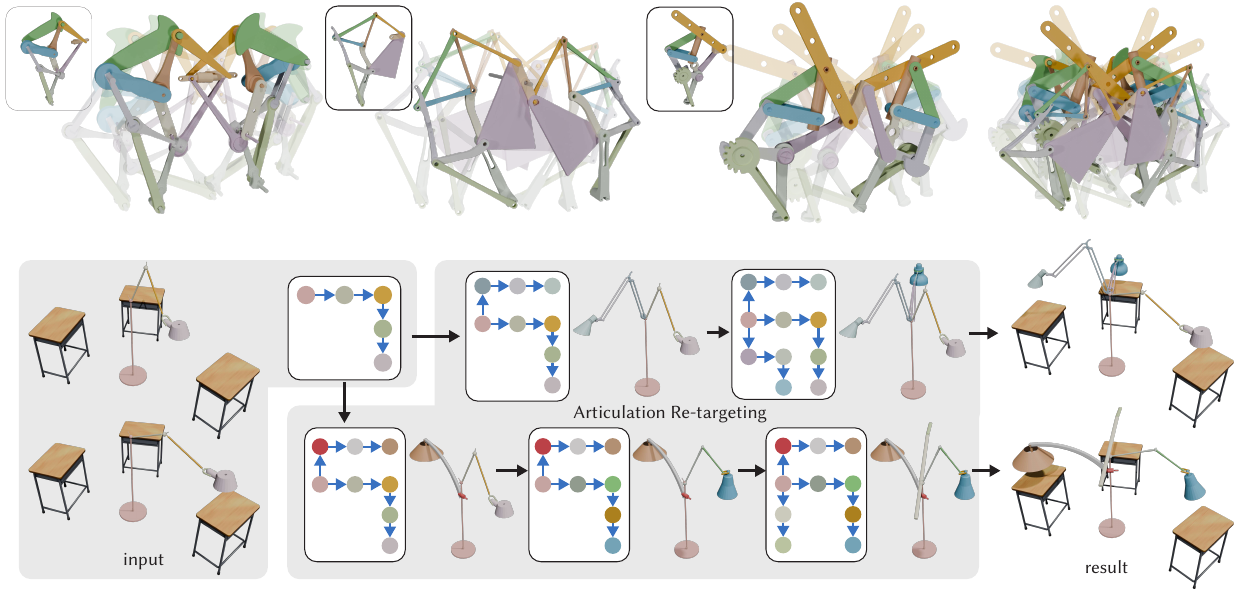}
    \end{center}
    \vspace{-2ex}
    \caption{\textbf{More results on kinematics-conditioned assembly and articulation re-targeting.} {Top:} Three geometric instantiations of Jansen's linkage using different retrieved parts, all preserving the same kinematic structure and characteristic gait. {Bottom:} Articulation re-targeting adapts a single-head lamp by proposing kinematic graph edits and re-instantiating each candidate, producing a multi-head lamp whose arms can reach the desired tabletop locations.}
    \label{fig:app_jansen}
\end{figure*}

\begin{figure*}[h]
    \begin{center}
    \includegraphics[width=1\linewidth]{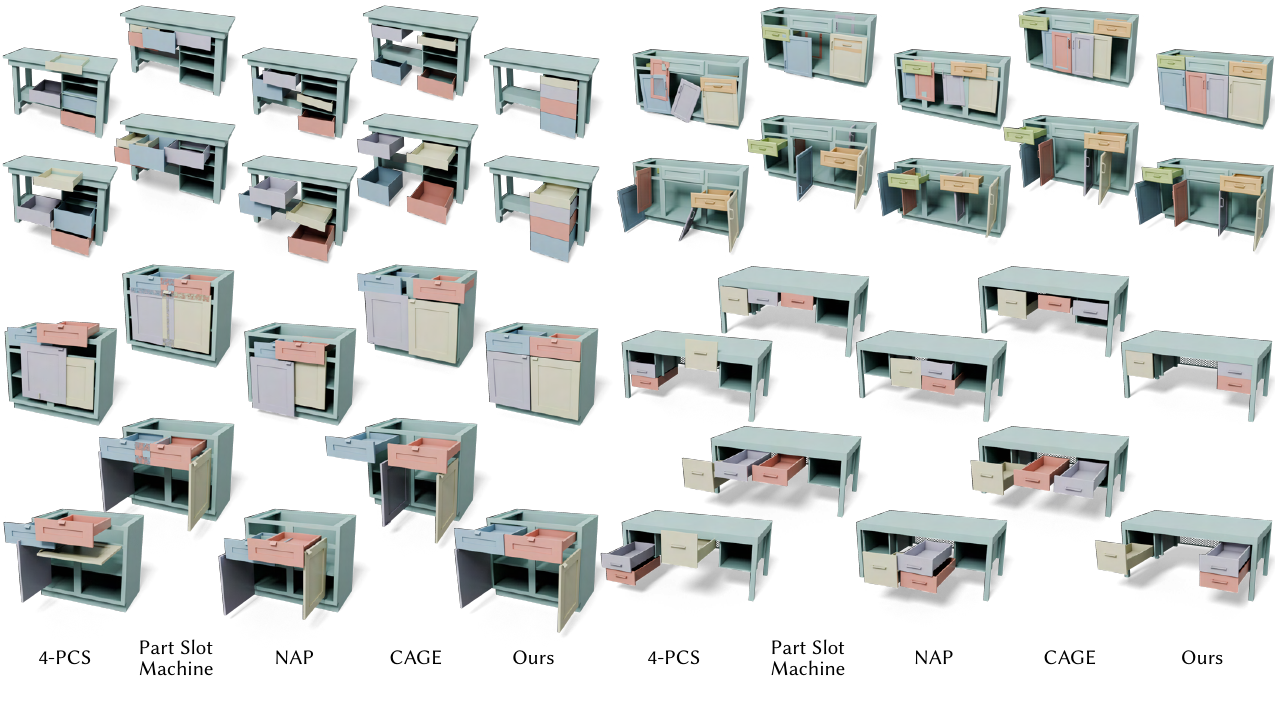}
    \end{center}
    \caption{\textbf{Qualitative comparison.} Geometry-only methods (4-PCS, Part Slot Machine) frequently misorient parts; Kinematics-aware baselines (NAP and CAGE) improve orientation but still yield floating or colliding parts. Our method aligns parts correctly and achieves fully functional articulation.}
    \label{fig:expr_baseline_2}
\end{figure*}

Our method can also serve as a lightweight post-processing stage for learning-based articulated generation.
As shown in Fig.~\ref{fig:post_processing}, starting from the result of an image-based generative model~\citep{FreeArt3D} that exhibits misaligned moving parts or incorrect attachments, we retrieve compatible parts and re-instantiate their kinematic attachments to recover clean door/drawer alignments.



\begin{table}[t]
  \caption{\textbf{Quantitative comparison with four baselines.} Our approach with kinematic-aware attachment delivers the highest scores across all geometric, kinematic, and functional metrics. 
  }
  \vspace{-2ex}
  \centering
  \setlength{\tabcolsep}{3pt}
  \renewcommand{\arraystretch}{1.1}
  \begin{tabular}{ccccccc}
    \toprule
         & \multicolumn{1}{c}{\textbf{Geometric}} 
         & \multicolumn{2}{c}{\textbf{Kinematic}} 
         & \multicolumn{2}{c}{\textbf{Functional}} \\[-0.2em]
    \cmidrule(lr){2-2} \cmidrule(lr){3-4} \cmidrule(lr){5-6}
    Method &
    Rooted$\uparrow$ &
    COV$\uparrow$ &
    MMD$\downarrow$ &
    Stable$\uparrow$ &
    AOR$\downarrow$ \\
    \midrule
    4-PCS               & 95.4\% & 56.9\% & 0.092  & 81.5\% & 16.4\% \\
    Part Slot Machine   & 98.5\% & 63.1\% & 0.083  & 84.6\% & 10.7\% \\
    \midrule
    
    NAP                 & 96.9\% & 69.2\% & 0.072  & 92.3\% & 3.5\%  \\
    CAGE                & \textbf{100\%}  & 73.8\% & 0.051  & 90.7\% & 1.1\%  \\
    \midrule
    
    Ours      & \textbf{100\%}  & \textbf{83.1\%} & \textbf{0.038}  & \textbf{96.9}\% & \textbf{0.1\%}  \\
    \bottomrule
  \end{tabular}
  \vspace{-2ex}
  \label{tab:metricWise}
\end{table}

\section{Discussion and Conclusion}
\label{sec:limitations}

We introduced kinematic kitbashing, an optimization framework that instantiates an abstract kinematic graph by reusing articulated parts from existing libraries. Our kinematics-aware VDF attachment preserves exemplar sockets across each joint's motion range, and we incorporate black-box task objectives through annealed Langevin sampling to support kinematic-conditioned assembly, functionality-guided synthesis, and re-targeting.
\fr{
The manual effort of our method lies in authoring the mechanism specification and curating the source kit, rather than manually arranging the articulated assembly. 
}

While the framework effectively enables functional assembly, it faces several computational and geometric challenges. 
\fr{
Our method promotes attachment compatibility by matching local attachment neighborhoods using the kinematics-aware VDF integrated over the sampled joint motion range. 
This is intended to preserve plausible attachments and clearances during articulation, rather than to guarantee fabrication-grade tolerances or full physical realism for downstream simulation or real-world use. 
Our method is intended as an offline optimization-based framework rather than a real-time system.
}
The annealed Langevin sampler requires a high number of iterations to achieve convergence, and runtime scales linearly with both the complexity of the kinematic graph and the number of articulation poses sampled for energy evaluation. 
Furthermore, the system does not provide hard guarantees for collision-free articulation because collision terms in $E_{\text{func}}$ are treated as soft penalties. 
This approach may occasionally overlook rare collisions occurring at unsampled poses.
The attachment prior is also fundamentally exemplar-driven, which can lead to failure cases when the target parent lacks a compatible neighborhood for the sourced part or when the library lacks candidates with suitable proportions. 

Several research avenues could significantly enhance the robustness and efficiency of this framework. 
\fr{
Our method does not currently propose entirely new kinematic graphs or high-level task specifications from scratch. More expressive graph editing, such as automatically exploring new subgraph topologies, would be a valuable extension.
}
Recent work on language-guided object--object alignment, such as Copy-Transform-Paste~\cite{gatenyo2026copytransformpaste}, suggests a complementary direction in which high-level textual intent and geometric contact constraints could help propose or rank candidate attachments before kinematics-aware refinement.
To accelerate convergence, parallel tempering-style schemes could be implemented to manage multiple Langevin chains at varying temperatures, allowing for more efficient state exchanges and exploration of the energy landscape~\cite{shih2023parallel}. 
Integrating explicit articulated collision-checking and resolution~\cite{JacobsonG16}
could provide hard guarantees. 




\bibliographystyle{ACM-Reference-Format}
\bibliography{sample-base}

\clearpage

\begin{figure*}[h]
    \begin{center}
    \includegraphics[width=1\linewidth]{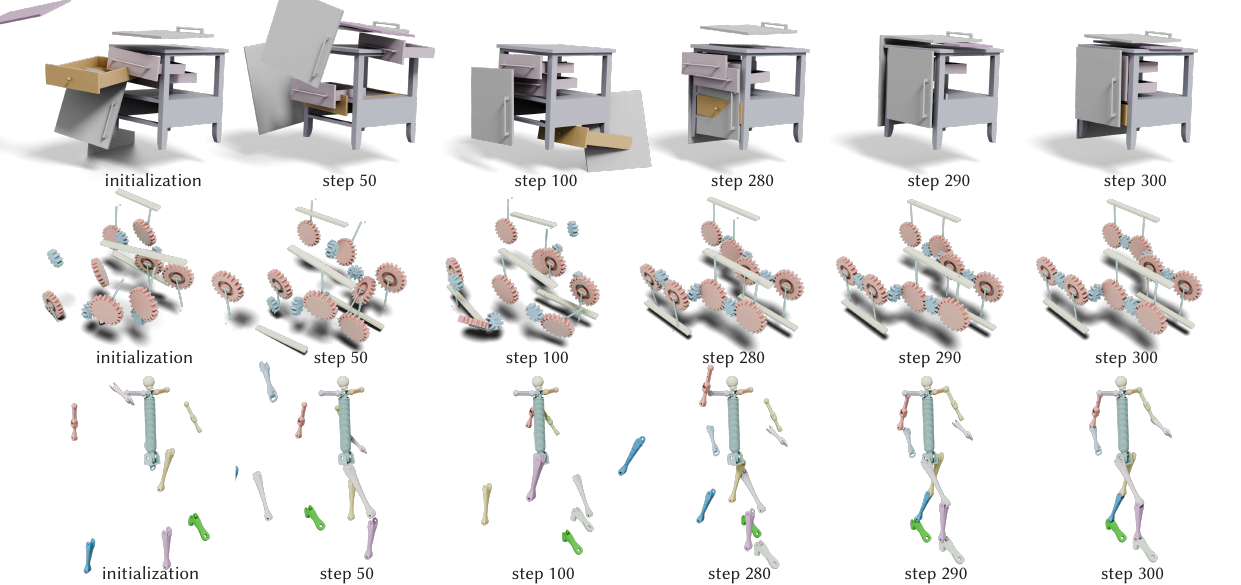}
    \end{center}
    \caption{\textbf{More intermediate results of annealed Langevin sampling.} Langevin dynamics run with both the kinematic attachment term and the functionality term gradually assemble the parts and end with a mechanism whose end-effectors reach the targets.}
    \label{fig:intermediate_sampling}
\end{figure*}

\clearpage

\newpage

\appendix
\providecommand{\artMesh}{A}            
\providecommand{\artShape}{S}          
\providecommand{\artPlacedMesh}{\tilde{A}} 
\providecommand{\artPlacedShape}{\tilde{S}} 
\providecommand{\jointT}{T}            
\providecommand{\jointq}{\boldsymbol{\theta}} 
\providecommand{\optPConcat}{P}        
\providecommand{\optP}{P}              
\providecommand{\mR}{R}                
\providecommand{\vd}{\mathbf{d}}       
\providecommand{\V}{V}                 
\providecommand{\mF}{F}                

\section{Supported Joint Types}
\label{app:joint-intro}
Our framework supports generic rigid-body joints.
Across the paper we use four standard joint types:
revolute (hinge, 1 DoF), prismatic (slider, 1 DoF), cylindrical (co-axial rotation and translation, 2 DoF), and Cartesian/planar sliders (independent translation along three orthogonal axes, 3 DoF). We use their conventional homogeneous-transform formulations; detailed expressions are available in \cite{featherstone2014rigid}.

\section{Detailed Formulations of Functionality Objectives}
\label{app:functionality-obj}

Below we give explicit mathematical formulations for the functionality terms used in our examples.
In practice, several of these terms are evaluated via an inverse kinematics (IK) solver, collision queries, or mesh processing.

We use $\optPConcat=\{(\mR_k,\vd_k,s_k)\}_{k=1}^K$ to denote the per-part similarity transforms that place the rest-pose parts in a shared world frame.
$\jointq\in\mathbb{R}^d$ stacks all joint coordinates (with limits $\jointq^{\text{lo}},\jointq^{\text{hi}}$), and $\jointT(\cdot;\jointq)$ denotes forward kinematics.
We write $\mathbf{p}(\optPConcat,\jointq)$ for any end-effector feature point in world coordinates obtained by applying $\optPConcat$ and then $\jointT$.

\subsection{Target reachability via inverse kinematics}
Let $H$ end-effectors (e.g., lamp heads) have world-space feature points
$\mathbf{p}_h(\optPConcat,\jointq)\in\mathbb{R}^3$ under joint parameters $\jointq$.
In our implementation, each $\mathbf{p}_h$ is the centroid of a fixed vertex subset on the corresponding end-effector mesh.
Given desired target points $\mathbf{t}_h\in\mathbb{R}^3$, we define a reachability objective as the optimal IK residual:
\begin{align}
E^{\mathrm{func}}_{\text{reach}}(\optPConcat)
&= \min_{\jointq\in[\jointq^{\text{lo}},\jointq^{\text{hi}}]}
\sum_{h=1}^{H}\big\|\mathbf{p}_h(\optPConcat,\jointq)-\mathbf{t}_h\big\|_2^2
\;+
\mu\,\Phi_{\text{limit}}(\jointq).
\label{eq:app_reach}
\end{align}
Here $[\jointq^{\text{lo}},\jointq^{\text{hi}}]$ stacks the per-joint limits.
We implement the joint-limit penalty as a squared hinge:
$[x]_+=\max(0,x)$ and
\begin{align}
\Phi_{\text{limit}}(\jointq)
=\sum_{j=1}^{d}\Big([\jointq_j-\jointq^{\text{hi}}_j]_+^2 + [\jointq^{\text{lo}}_j-\jointq_j]_+^2\Big).
\end{align}
We set $\mu=10^2$ in all reachability experiments and evaluate Eq.~\eqref{eq:app_reach} with a standard IK solver.

We use this reachability objective in the multi-head lamp target reaching (Fig.~7 and Fig.~14) and the lamp re-targeting (Fig.~12, bottom).

\subsection{Trajectory tracking}
Given a desired end-effector trajectory $\{\mathbf{q}_i\}_{i=1}^{N}$ in world coordinates,
we evaluate the best achievable tracking error over a sequence of joint states $\{\jointq_i\}$:
\begin{align}
E^{\mathrm{func}}_{\text{track}}(\optPConcat)
&= \min_{\{\jointq_i\}}\;\sum_{i=1}^{N}\big\|\mathbf{p}(\optPConcat,\jointq_i)-\mathbf{q}_i\big\|_2^2
\;+
\eta\sum_{i=2}^{N}\big\|\jointq_i-\jointq_{i-1}\big\|_2^2
\\
&\text{s.t.}\quad \jointq_i\in[\jointq^{\text{lo}},\jointq^{\text{hi}}]\;\;\forall i.
\label{eq:app_track}
\end{align}
The optional smoothness term discourages discontinuous motion when optimizing a full pose sequence;
we set $\eta=10^{-2}$ in all tracking experiments.

We employ this objective in the gear-driven paddle trajectory tracking (Fig.~8, left) and the mechanical humanoid (Fig.~8, right).

\subsection{Kinematic synchronization}
To couple two scalar joint coordinates $\theta_a$ and $\theta_b$ (extracted from $\jointq$), we use
\begin{align}
E^{\mathrm{func}}_{\text{sync}}(\optPConcat)
&= \min_{\{\jointq_i\}}\;\sum_{i=1}^{N}\Big(\theta_a(\jointq_i)+\rho\,\theta_b(\jointq_i)-\theta_0\Big)^2,
\label{eq:app_sync}
\end{align}
where $\theta_a(\jointq)$ and $\theta_b(\jointq)$ select the corresponding scalar coordinates from $\jointq$,
$\rho$ is the target ratio, and $\theta_0$ is a phase offset. In our gear examples we use $\rho=1$ and $\theta_0=0$.

\noindent\emph{Used in.}
Counter-rotating paddle/gear synchronization (Fig.~8, left).

\subsection{Collision and swept-volume penalties}We penalize inter-penetrations either between (i) pairs of parts or (ii) parts and scene obstacles.
Let $\mathcal{P}$ be a set of mesh pairs to test.
For pose $i$, define the placed mesh for part $k$ as $\artPlacedMesh^{(i)}_k(\optPConcat)$.
Let $d_{B}(\mathbf{x})$ be the signed distance from point $\mathbf{x}$ to mesh $B$ (positive outside, negative inside).
With clearance margin $\delta\ge 0$, define a soft barrier
\begin{align}
\phi_{\delta}(d)=\max(0,\delta-d)^2.
\end{align}
We use the sampled-point collision penalty
\begin{align}
E^{\mathrm{func}}_{\text{coll}}(\optPConcat)
&= \sum_{i=1}^{N}\;\sum_{(k,\ell)\in\mathcal{P}}
\frac{1}{|\mathcal{X}_{k,i}|}\sum_{\mathbf{x}\in\mathcal{X}_{k,i}}
\phi_{\delta}\Big(d_{\artPlacedMesh^{(i)}_{\ell}(\optPConcat)}(\mathbf{x})\Big),
\label{eq:app_collision}
\end{align}
where $\mathcal{X}_{k,i}$ is a set of sampled surface points on part $k$ at pose $i$ (we use uniformly sampled mesh vertices),
and $\delta$ specifies the desired clearance; we use $\delta=0$ (penetration-only) unless otherwise stated.
For swept-volume penalties, we discretize the motion cycle with multiple poses $\{\jointq_i\}$; Eq.~\eqref{eq:app_collision} already accumulates across poses.

This penalty is used in the collision-free actuation for functionality-guided assemblies (Fig.~8) and for the cabinet re-targeting in a cluttered scene (Fig.~9).

\subsection{Packing / fit-in-box constraints}
To pack a mechanism into a target axis-aligned box centered at $\mathbf{c}$ with half-sizes $\mathbf{h}$,
we penalize points outside the box.
For a point $\mathbf{x}$, define componentwise violation $\mathbf{v}(\mathbf{x})=|\mathbf{x}-\mathbf{c}|-\mathbf{h}$.
Then
\begin{align}
E^{\mathrm{func}}_{\text{box}}(\optPConcat)
&= \sum_{\mathbf{x}\in\mathcal{X}}\big\|\max(\mathbf{0},\mathbf{v}(\mathbf{x}))\big\|_2^2,
\label{eq:app_box}
\end{align}
where $\max(\cdot,\cdot)$ is applied componentwise and $\mathcal{X}$ are sampled points on the object in the designated packed configuration (typically a single pose).

We demonstrate the application of this term in the folding workstation packing into a bounding box (Fig.~8, bottom).

\subsection{Closed-state coverage}
For cabinet re-targeting we encourage a closed pose to cover the cabinet opening region.
Let $\Pi(\cdot)$ be an orthographic projection onto the cabinet front plane.
Let $\Omega$ be the target opening region on that plane.
For the set of door panels $\mathcal{D}$ in the closed configuration, define the union of projected door regions
$D(\optPConcat)=\bigcup_{k\in\mathcal{D}}\Pi(\artPlacedMesh^{(\text{closed})}_k(\optPConcat))$.
Then
\begin{align}
E^{\mathrm{func}}_{\text{cover}}(\optPConcat)
&=1-\frac{\operatorname{area}(\Omega\cap D(\optPConcat))}{\operatorname{area}(\Omega)}.
\label{eq:app_cover}
\end{align}
In the cabinet scenes we combine coverage with collision penalties to ensure collision-free actuation (see Fig.~9).

\section{Robust Kinematic Attachment using Vector Distance Fields}
\label{app:attachment-robust}

To enhance the robustness of kinematic attachment, we follow the methods in~\cite{sorkine2017least, zhang2021fast} and use the point-to-plane residual with Welsch's function rather than $l_2$ norm:
\begin{equation}
\begin{gathered}
    \Ekm_{k,i} = \sum_i\psi_\nu \Big(\mathbf{r}_k^{(i)}\cdot\mathbf{\hat{n}}_k^{(i)}\Big), \\ \mathbf{r}_k^{(i)}=\mR_k \vdf\Big(\mathcal{A}_{k}^{(i)};\MParent_k\Big)\!-\!\vdf\Big(\artPlacedMesh^{(i)}_k(\optPConcat); \MParentNew_k\Big)\,
\end{gathered}
\end{equation}
where $\mathbf{\hat{n}}_k^{(i)}$ is obtained by concatenating all the surface normals of the new parent mesh $\MParentNew_k$ at every cloest point corresponding to the verties of the transformed part $\artPlacedMesh^{(i)}_k(\optPConcat)$.
$\psi_\nu$ is the Welsch's function:
\begin{align}
    \psi_\nu(x) = 1 - \exp(-\frac{x^2}{2\nu^2}).
\end{align}
We choose $\nu\!=\!0.5$.

\section{Formulation of Langevin Dynamics}

The task objective $\Efunc$ encompasses a diverse range of evaluations,  
such as inverse kinematics and swept-volume contact checks, so using gradient-based solvers would require bespoke derivations for each new target.
Meanwhile, the optimization variable $\optPConcat$ spans $6K$ continuous DoFs, already larger than $30$ for a six-part assembly, where evolutionary or simulated-annealing schemes lose efficiency \cite{knight2007reducing, zhou2024evolutionary}. 
%
We address both challenges with annealed Langevin dynamics.
Treating the combined objective of the kinematic attachment term $\Ekm$ and the functionality term $\Efunc$ as an unnormalized probability density allows us to replace deterministic minimization with stochastic sampling.
The sampling process converges towards high-probability \emph{modes}, i.e., configurations with concentrated low objective values, while naturally discarding improbable outliers, so optimal placements can be found without the need for analytic gradients.
A detailed analysis of Langevin dynamics as a robust mode finder, and its relation to mean-shift and voting schemes, is discussed in~\cite{je2024robust}.



Specifically, Langevin dynamics samples a sequence of rigid transform sets $\{\optPConcat^s|s = S,...,0\}$.
It starts with $\optPConcat^S$ drawn from Gaussian noises and iteratively steers towards the target distribution 
\begin{align}
    p_0(\optPConcat^0) \propto \exp\bigl(-\frac{\Ekm(\artPlacedShape(\optPConcat^0))}{\lambda}\bigr) \exp(-\Efunc(\artPlacedShape (\optPConcat^0))).
\end{align}
The update at each step follows
\begin{align}
    \optPConcat^{s-1} \leftarrow \optPConcat^s +  \frac{\alpha_s}{2}\nabla_{\optPConcat^s}\log p_{s}(\optPConcat^s) + \sqrt{\alpha_s}\epsilon, \ s = S,...,1 \,,
\end{align}
where $\alpha_s$ is the step size, $\epsilon\!\sim\!\mathcal{N}(0,\mathbf{I})$ is the noise.
The \emph{score function} $\nabla_{\optPConcat^s}\log p_{s}(\optPConcat^s)$ is estimated
via Monte Carlo sampling: 
\begin{align*}
\nabla_{\optPConcat^s} \log p_s(\optPConcat^s) 
&=\frac{\sum_{\optPConcat^0\in \mathscr{P}^0} p_0(\optPConcat^0)\nabla_{\optPConcat^s} \log p_{s|0}(\optPConcat^s|\optPConcat^0)}{\sum_{\optPConcat^0\in \mathscr{P}^0} p_0(\optPConcat^0)},
\end{align*}
where 
$\mathscr{P}^0$ is a batch of samples drawn from the proposal function 
$p_{s|0}(\optPConcat^s|\optPConcat^0)$.
It gives an unbiased approximation 
~\cite{song2019generative, pan2024model}.
A full derivation is provided in the Appendix. 

We run Langevin dynamics for $S\!=\!300$ iterations.
To evaluate $p_0(\cdot)$, we first apply the kinematics-aware attachment optimization and then compute $\Ekm$ and $\Efunc$ on the resulting configuration.
The score estimation uses automatic differentiation with $30$ samples.

\section{Formulation of the Score Function}
\label{app:score-function}
Following Bayes' rule, the score function $\nabla_{\optPConcat^s}\log p_{s}(\optPConcat^s)$ can be estimated using Monte Carlo sampling:
\begin{align*}
\nabla_{\optPConcat^s} \log p_s(\optPConcat^s) 
&= \frac{\nabla_{\optPConcat^s} \int p_{s|0}(\optPConcat^s|\optPConcat^0)p_0(\optPConcat^0)\mathrm{d}\optPConcat^0}{\int p_{s|0}(\optPConcat^s|\optPConcat^0)p_0(\optPConcat^0)\mathrm{d}\optPConcat^0} \\
&= \frac{\int \nabla_{\optPConcat^s} p_{s|0}(\optPConcat^s|\optPConcat^0)p_0(\optPConcat^0)\mathrm{d}\optPConcat^0}{\int p_{s|0}(\optPConcat^s|\optPConcat^0)p_0(\optPConcat^0)\mathrm{d}\optPConcat^0} \\
&= \frac{\int p_{s|0}(\optPConcat^s|\optPConcat^0) p_0(\optPConcat^0) \nabla_{\optPConcat^s} \log p_{s|0}(\optPConcat^s|\optPConcat^0) \mathrm{d}\optPConcat^0}{\int p_{s|0}(\optPConcat^s|\optPConcat^0)p_0(\optPConcat^0)\mathrm{d}\optPConcat^0} \\
&=\frac{\sum_{\optPConcat^0\in \mathscr{P}^0} p_0(\optPConcat^0)\nabla_{\optPConcat^s} \log p_{s|0}(\optPConcat^s|\optPConcat^0)}{\sum_{\optPConcat^0\in \mathscr{P}^0} p_0(\optPConcat^0)},
\end{align*}
where $\mathscr{P}^0$ denotes the set of samples drawn from the proposal function $p_{s|0}(\optPConcat^s|\optPConcat^0)$, which is the transition density kernel of Langevin dynamics.

Since $\optPConcat^s=\{(\mR_k^s,\vd_k^s,s_k^s)\}_{k=1}^K$ lies on the product Lie group $\mathrm{Sim}(3)^K$, the transition kernel
$p_{s\mid 0}(\optPConcat^s \mid \optPConcat^0)$ cannot be expressed as a simple Gaussian in Euclidean coordinates.
We therefore factor the update into Euclidean and rotational components.
Translations and (log-)scales are evolved with standard Euclidean Langevin dynamics, while rotations are updated using an isotropic Gaussian kernel on $\mathrm{SO}(3)$~\cite{nikolayev1997normal}.
This product construction yields Langevin dynamics that remain on $\mathrm{Sim}(3)^K$.

\section{More Details of Kinematics-Conditioned Assembly with Part Retrieval}
\label{app:retrieval}

This section expands Sec.~4.1 (retrieval setting), where parts are not provided explicitly and are instead retrieved from a large library.
The goal is to instantiate a kinematic graph whose nodes carry coarse geometric and joint cues.

\noindent\emph{Notation.}
We use $v$ to index nodes of the target kinematic graph and $\ell$ to index candidate parts in the library.
For each node $v$ we assume an oriented bounding box (OBB) frame $\mathcal{B}_v$, and we store geometric/joint cues in this local coordinate system.

\subsection{Node descriptors and canonical frames}
For each graph node $v$, we assume an oriented bounding box (OBB) frame $\mathcal{B}_v$ is given (or inferred from a proxy).
In this local frame we store a descriptor
\begin{align}
\mathcal{D}(v)=\big(\mathbf{b}_v,\;\tau_v,\;\mathbf{c}_v,\;\hat{\mathbf{a}}_v\big),
\end{align}
where $\mathbf{b}_v\in\mathbb{R}_+^3$ are OBB side lengths,
$\tau_v$ is the joint type,
$\mathbf{c}_v$ is a joint anchor (pivot) point, and $\hat{\mathbf{a}}_v$ is a unit joint axis (when defined).
We store $\mathbf{c}_v$ in {normalized} OBB coordinates (dividing each component by $\mathbf{b}_v$), so that anchor comparisons are scale-invariant.
For every candidate library part $\ell$, we precompute the same descriptor $\mathcal{D}(\ell)$ in its own OBB frame.

\subsection{Retrieval distance and scale adaptation}
We retrieve parts by a weighted distance
\begin{align}
D(v,\ell)
&=w_{\text{box}}\,\big\|\log \mathbf{b}_v-\log \mathbf{b}_{\ell}\big\|_2^2
+w_{\text{anc}}\,\big\|\mathbf{c}_v-\mathbf{c}_{\ell}\big\|_2^2
+w_{\text{axis}}\,\Big(1-\big|\hat{\mathbf{a}}_v^{\top}\hat{\mathbf{a}}_{\ell}\big|\Big),
\label{eq:app_retrieval_dist}
\end{align}
evaluated only among candidates with matching joint type ($\tau_v=\tau_{\ell}$).
The $\log(\cdot)$ is applied componentwise to the side-length vectors.
The axis term is applied only when the joint type has a well-defined axis (revolute/prismatic/cylindrical); otherwise it is set to $0$.
All three terms are dimensionless under our normalization, and we set $w_{\text{box}}=w_{\text{anc}}=w_{\text{axis}}=1$ in experiments.
The log-size term encourages similarity up to uniform scale, consistent with our similarity-transform optimization over $s_k$.
After retrieval, the final scale is determined by minimizing the attachment energy across the kinematic graph, so retrieved parts can be resized to fit (see Fig.~5).

\subsection{Selecting a full part set}
Let $\mathcal{L}(v)$ be the top-$M$ retrieved candidates for node $v$.
A full instantiation is an assignment $\mathcal{A}:V\rightarrow\bigcup_v\mathcal{L}(v)$ choosing one part per node.
We generate a small set of promising assignments using beam search over nodes
(scoring partial assignments by the sum of $D(v,\ell)$ and optionally penalizing repeated source assets for diversity),
then run the continuous kinematics-conditioned assembly optimization (minimizing $E_{\text{km}}$) to finalize the transforms $\optPConcat$.
In all retrieval experiments we use $M=50$ candidates per node and a beam width of $20$.

\section{Detailed Formulation of Kinematic Graph Editing}
\label{app:graph_editing}

Articulation retargeting (Sec.~4.3) augments continuous placement optimization with discrete edits to the kinematic graph.
We cast this as a discrete--continuous (bilevel) optimization problem:
\begin{align}
\min_{\mathcal{G}\in\mathcal{N}(\mathcal{G}_0)}\;
\Bigg[
\min_{\optPConcat}\;
\Big(
E^{\text{km}}\big(\artPlacedShape(\optPConcat;\mathcal{G})\big)
+\lambda\,E^{\text{func}}\big(\artPlacedShape(\optPConcat;\mathcal{G})\big)
\Big)
\Bigg]
+\gamma\,C_{\text{edit}}(\mathcal{G},\mathcal{G}_0),
\label{eq:app_graph_bilevel}
\end{align}
where $\mathcal{G}_0$ is the initial graph, $\mathcal{N}(\mathcal{G}_0)$ denotes a neighborhood of graphs reachable from $\mathcal{G}_0$ via a small number of local edits, and $C_{\text{edit}}$ penalizes excessive deviation from the original schema.
For a fixed edited graph $\mathcal{G}$, the inner problem is identical to the functionality-guided placement optimization, except that forward kinematics and parent--child attachment are evaluated using the edited connectivity.
We follow the discrete proposal-and-selection strategy of~\citet{DesignForDecent}.

We consider two local graph rewrite operators that preserve the high-level semantic intent:
(i) \emph{Edge splitting} inserts an intermediate link by replacing an edge $(u\!\rightarrow\!v)$ with $(u\!\rightarrow\!w\!\rightarrow\!v)$, where $w$ is a newly introduced node whose part is retrieved (Sec.~\ref{app:retrieval}); and
(ii) \emph{Node replacement / joint-variant swap} re-retrieves the part assigned to an existing node from a constrained candidate set, enabling alternatives such as hinge-side changes or joint-placement variants.
Each operator incurs a unit edit cost, and $C_{\text{edit}}(\mathcal{G},\mathcal{G}_0)$ is defined as the total cost of the edits applied to transform $\mathcal{G}_0$ into $\mathcal{G}$.

\section{Implementation Details}
\label{app:implement-details}
Our experiments run on a desktop with an AMD Ryzen 9 5950X (16 cores, 64 GB RAM) and an NVIDIA RTX 8000 GPU.
For each part-parent pair, we pre-compute $5$ articulation snapshots ($T\!=\!5$); a complete kinematic-attachment solve across these snapshots takes about $4$~s.
Annealed Langevin dynamics runs for 300 iterations, and each iteration invokes one kinematic-attachment update, giving a total wall-clock time of roughly $20$~min per assembly.

We reuse kinematic parts from three sources: (i) PartNet-Mobility, which includes everyday articulated objects~\cite{PartNetMobility}, (ii) JoinABLe, which is a large library of CAD-style jointed parts~\cite{Joinable}, and (iii) artist-curated assets, which is a small collection of additional articulated models (including commissioned designs and purchased models, e.g., Blender Market).

\end{document}